\documentclass[10pt,twocolumn,letterpaper]{article}

\usepackage{wacv}
\usepackage{times}
\usepackage{epsfig}
\usepackage{graphicx}
\usepackage{amsmath}
\usepackage{amssymb}

\usepackage{physics}
\usepackage[percent]{overpic}
\usepackage{subcaption}

\makeatletter
\newlength\mylen
\DeclareRobustCommand*\mytextsuperscript[1]{%
	\@Textsuperscript{\selectfont#1}}
\def\@Textsuperscript#1{%
	\settoheight\mylen{\fontsize\f@size\z@ A}%
	{\m@th\ensuremath{\raise.5\mylen\hbox{\fontsize\sf@size\z@#1}}}}
\makeatother

\def\wacvPaperID{0609} % Enter the WACV Paper ID here

\wacvfinalcopy % *** Uncomment this line for the final submission

\ifwacvfinal
\fi

\ifwacvfinal
\usepackage[breaklinks=true,bookmarks=false]{hyperref}
\else
\usepackage[pagebackref=true,breaklinks=true,colorlinks,bookmarks=false]{hyperref}
\fi

\begin{document}

%%%%%%%%% TITLE
\title{Illumination Normalization by Partially Impossible \\ Encoder-Decoder Cost Function}

% \author{First Author\\
% Institution1\\
% Institution1 address\\
% {\tt\small firstauthor@i1.org}
% % For a paper whose authors are all at the same institution,
% % omit the following lines up until the closing ``}''.
% % Additional authors and addresses can be added with ``\and'',
% % just like the second author.
% % To save space, use either the email address or home page, not both
% \and
% Second Author\\
% Institution2\\
% First line of institution2 address\\
% {\tt\small secondauthor@i2.org}
% }

\author{Steve Dias Da Cruz\,\mytextsuperscript{1,2,3}\\
	{\tt\small steve.dias-da-cruz@iee.lu}
	\and
	Bertram Taetz\,\mytextsuperscript{3}\\
	{\tt\small bertram.taetz@dfki.de}
	\and
	Thomas Stifter\,\mytextsuperscript{1}\\
	{\tt\small thomas.stifter@iee.lu}
	\and
	Didier Stricker\,\mytextsuperscript{2,3}\\
	{\tt\small didier.stricker@dfki.de} \vspace{0.2cm} \\
	\mytextsuperscript{1}\,IEE S.A. \hspace{0.5cm} \mytextsuperscript{2}\,University of Kaiserslautern \hspace{0.5cm} \mytextsuperscript{3}\,German Research Center for Artificial Intelligence
}

\maketitle
\thispagestyle{empty}

%%%%%%%%% ABSTRACT
\begin{abstract}
Images recorded during the lifetime of computer vision based systems undergo a wide range of illumination and environmental conditions affecting the reliability of previously trained machine learning models. Image normalization is hence a valuable preprocessing component to enhance the models' robustness. To this end, we introduce a new strategy for the cost function formulation of encoder-decoder networks to average out all the unimportant information in the input images (\eg environmental features and illumination changes) to focus on the reconstruction of the salient features (\eg class instances). Our method exploits the availability of identical sceneries under different illumination and environmental conditions for which we formulate a partially impossible reconstruction target: the input image will not convey enough information to reconstruct the target in its entirety. Its applicability is assessed on three publicly available datasets. We combine the triplet loss as a regularizer in the latent space representation and a nearest neighbour search to improve the generalization to unseen illuminations and class instances. The importance of the aforementioned post-processing is highlighted on an automotive application. To this end, we release a synthetic dataset of sceneries from three different passenger compartments where each scenery is rendered under ten different illumination and environmental conditions: \url{https://sviro.kl.dfki.de}
\end{abstract}

%%%%%%%%% BODY TEXT
\section{Introduction}

\begin{figure}
	\centering
	\begin{overpic}[width=0.31\linewidth]{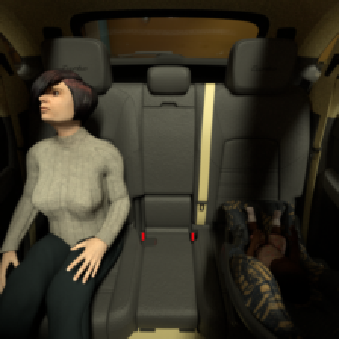}
		\put(7,33){\rotatebox{90}{\textcolor{white}{Input}}}
	\end{overpic}
	\includegraphics[width=0.31\linewidth]{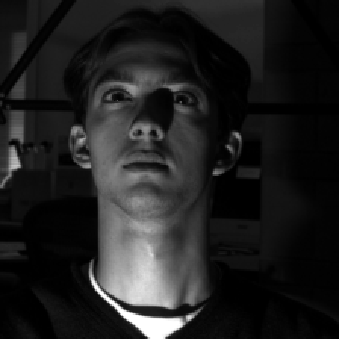}
	\includegraphics[width=0.31\linewidth]{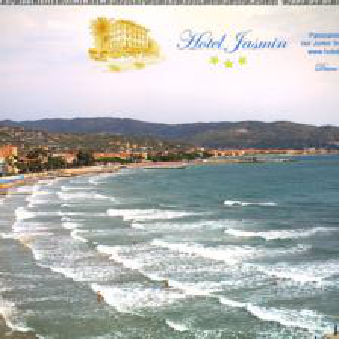}
	\vskip 0.1 \baselineskip
	\begin{overpic}[width=0.31\linewidth]{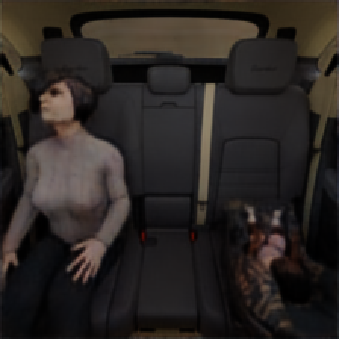}
		\put(7,25){\rotatebox{90}{\textcolor{white}{Cleaned}}}
	\end{overpic}
	\includegraphics[width=0.31\linewidth]{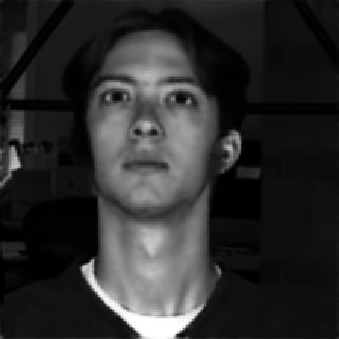}
	\includegraphics[width=0.31\linewidth]{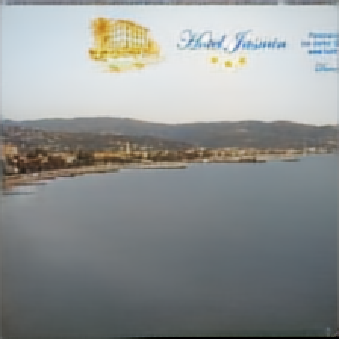}
	\caption{Results for the introduced partially impossible cost function. The input to the encoder-decoder model (first row) is transformed such that illumination and environmental features are averaged out and removed (second row).}
	\label{fig:teaser}
\end{figure}

Recording a sufficient amount of images to train and evaluate computer vision algorithms is usually a time consuming and expensive challenge. This is aggravated when the acquisition of images under various lightning and weather conditions needs to be considered as well. Notwithstanding the aforementioned data collection challenges, the performance of many machine learning algorithms suffer from changing illumination or environmental conditions, \eg SLAM \cite{glover2010fab}, place recognition \cite{olid2018single}, localization and classification \cite{maddern2014illumination}, semantic segmentation \cite{alshammari2018impact}, 3D human pose estimation \cite{robertini2018illumination} and facial expression recognition \cite{ruiz2018deep}. Since it is impracticable to wait for different weather conditions, day times and seasons to record images under as many variations as possible, it would be beneficial to train machine learning models to become invariant with respect to illumination and the exterior environment. Particularly for safety critical applications, as is common in the automotive industry, it would be of interest to reduce the amount of different illumination conditions necessary to guarantee reliable inference of machine learning models. Improvements on the aforementioned invariances would reduce the amount of mileage and images needed to be recorded and hence reduce the financial risk and time investment while improving the overall robustness of the deployed system. 

We aim to transform the input image by removing illumination and environmental features instead of computing more robust and invariant feature descriptors like SIFT \cite{lowe1999object} or enforcing illumination invariance in deep neural networks through data augmentation. We achieve this by exploiting the availability of sceneries under different illumination and/or environmental conditions. We will introduce a partially impossible reconstruction loss in Section \ref{sec:loss} which enforces similarity in the latent space of encoder-decoder models implicitly, in opposition to an explicit constraint \cite{antelmi2019sparse, zhang2015learning}. In contrast to shadow removal \cite{wang2018stacked, qu2017deshadownet} or relighting \cite{sun2019single,zhang2020portrait}, our method removes all the illumination and environmental features together. Our method is neither limited to a specific application where prior knowledge, \eg about faces \cite{zhang2020portrait, shu2017portrait}, needs to be included, nor does it need shadow and shadow-free image pairs \cite{wang2018stacked, qu2017deshadownet} to define a ground truth target. We highlight its applicability on multiple datasets and provide evidence for the usefulness of collecting images under these more challenging conditions. Example results on multiple datasets are shown in Fig. \ref{fig:teaser}. 

In this work, we focus on the automotive application of occupant classification in the vehicle interior rear bench to demonstrate our proposed method's applicability. To this end, we release a synthetic dataset for occupant classification in three vehicle interiors where each scenery is rendered under ten different illumination and environmental conditions. We will demonstrate the benefits of combining an encoder-decoder based approach for illumination and environmental feature removal together with a triplet loss regularizer in the latent space. The latter improves the nearest neighbour search on test samples and hence the reliability and generalization to unseen samples. We quantitatively assess this improvement based on the classification accuracy. Our key contributions can be summarized as follows:
\begin{itemize}
\item We introduce a partially impossible reconstruction cost function in encoder-decoder models to remove illumination and environmental features,
\item We highlight the importance of a triplet loss regularizer in the latent space of encoder-decoder models to improve generalization to unseen sceneries,
\item We release the SVIRO-Illumination dataset, which contains 1500 sceneries (once with people only and once with child and infant seats) from three vehicle interiors, where each scene is rendered under 10 different illumination and environmental conditions.
\end{itemize}

\section{Related Work}

\noindent\textbf{Datasets}: Recording identical, or similar, sceneries under different lightning or environmental conditions is a challenging task.	 Large scale datasets for identical sceneries under different lightning conditions are currently scarce. The Deep Portrait Relighting Dataset \cite{zhou2019deep} is based on the CelebA-HQ \cite{karras2018progressive} dataset and contains human faces under different illumination conditions. However, the re-illumination has been added synthetically. Regarding the latter constraint, we instead used the Extended Yale Face Database B \cite{GeBeKr01}, which is a dataset of real human faces with real illumination changes. While cross-seasons correspondence datasets prepared according to \cite{larsson2019cross} and based on RobotCar \cite{RobotCarDatasetIJRR} and CMU Visual Localization \cite{badino2011visual} could potentially be used for our investigation, the correspondences are usually not exact enough to have an identical scene under different conditions. Moreover, dominantly visible changing vehicles on the streets induce a large difference in the images. Alternative datasets such as St. Lucia Multiple Times of Day \cite{Glover2010ICRA} and Nordland \cite{olid2018single} suffer from similar problems. However, these datasets stem from the image correspondence search, place recognition and SLAM community. We adopt the Webcam Clip Art \cite{lalonde-siggraph-asia-09} to include a dataset for the exterior environment with changing seasons and day times instead. The latter contains webcam images of outdoor regions from different places all over the world. 

\noindent\textbf{Consistency in latent space}: Existing encoder-decoder based methods try to represent the information from multiple domains \cite{antelmi2019sparse} or real-synthetic image-pairs \cite{zhang2015learning} identically in the latent space by enforcing some similarity constraints, \eg the latent vectors should be close together. However, these approaches often force networks to reconstruct some (or all) of the images correctly in the decoder part. Forcing an encoder-decoder to represent two images (\eg same scenery, but different lightning) identically in the latent space, yet simultaneously forcing it to reconstruct both input images correctly implies an impossibility: The decoder cannot reconstruct two different images using the same latent space. Antelmi \etal \cite{antelmi2019sparse} adopted a different encoder-decoder for each domain, but as illumination changes are continuous and not discrete, we cannot have a separate encoder or decoder for each possible illumination.  

\noindent\textbf{Shadow removal and relighting}: Recent advances in portrait shadow manipulation \cite{zhang2020portrait} try to remove shadows cast by external objects and to soften shadows cast by the facial features of the subjects. While the proposed method can generalize to images taken in the wild, it has problems for detailed shadows and it assumes that shadows either belong to foreign or facial features. Most importantly, it assumes facial images as input and exploits the detection of facial landmarks and their symmetries to remove the shadows. Other shadow removal methods \cite{wang2018stacked, qu2017deshadownet} are limited to simpler images. The backgrounds and illumination are usually quite uniform and they contain a single connected shadow. Moreover, the availability of shadow and shadow-free image pairs provides the means of a well defined ground truth. However, this is is not possible for more complex scenes and illumination conditions for which a ground truth is not available or even impossible to define. Image relighting \cite{sun2019single, zhou2019deep} could potentially be used to change the illumination of an image to some uniform illumination. However, as noted in \cite{sun2019single,zhang2020portrait} relighting struggles with foreign or harsh shadows. While it is possible to fit a face to a reference image \cite{shu2017portrait}, this option is limited to facial images as well. 

\section{Method}
We will introduce our proposed partially impossible cost function for encoder-decoder networks to exploit the availability of identical sceneries under different lightning conditions. We will suggest to extend our method by applying a triplet loss regularizer in the latent space to improve generalization. This induces some useful properties such that more robust and reliable results on unseen test samples can be achieved by adopting the nearest neighbour search.

\subsection{Partially impossible reconstruction loss}
\label{sec:loss}

Our proposed partially impossible reconstruction cost function can be applied to any encoder-decoder neural network architecture. Instead of considering the standard autoencoder reconstruction loss defined as the difference between the input image and the decoder reconstruction, we formulate an alternative reconstruction loss based on the decoder reconstruction and a new variation of the input image. 

\begin{figure}
	\centering
	\includegraphics[width= \linewidth]{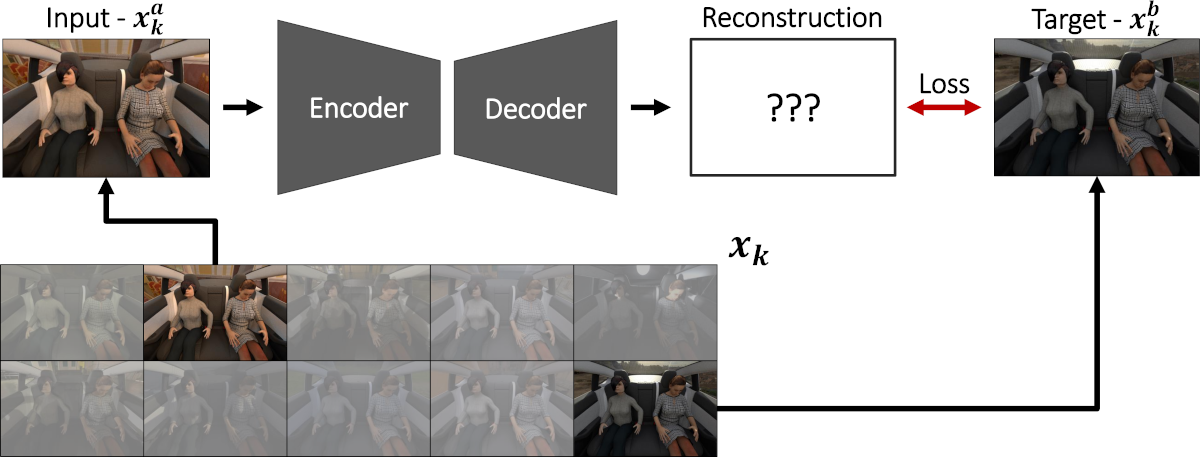}
	\caption{For each scenery $x_k$, we randomly select two images $x_k^a$ and $x_k^b$ under different lightning and environmental conditions. One is considered as input $x_k^a$ to the encoder, the other one as target image $x_k^b$ for the reconstruction.}
	\label{fig:method}
\end{figure}

Let $\mathcal{X}$ be the set of all training images and $x_k$ be the $k$th scene of the training data. For each scene we have $n$ images, where each image represents the same scene under different lightning and/or environmental conditions. We denote by $x_k^j$ the $j$th image out of the $n$ images for scene $k$. Hence, the training data can be expressed as $x_k^j \in \mathcal{X}$ for $k \in[0, N]$ and $j \in [0, n]$, where $N$ is the total number of unique scenes. Moreover, $x_k^j \in x_k$ for $j \in [0, n]$. Denote by $\mathcal{X}_m \subset \mathcal{X}$ a subset containing $m$ number of sceneries from all the sceneries available in the training data.  During training, the batches iterate over the $x_k$ and for each $x_k$ we randomly select $a,b \in [0, n], a \neq b$ to get $x_k^a, x_k^b \in x_k$. Finally, $x_k^a$ is considered input to the encoder-decoder network and $x_k^b$ is considered as the target for the reconstruction loss. The aforementioned method is illustrated in Fig. \ref{fig:method}. The reconstruction loss can hence be formulated as
\begin{equation}
	 \mathcal{L}_R(\mathcal{X}_m; \theta, \phi) = \sum_{k=0}^{m} \mathrm{r}\left(\mathrm{h}_{\theta}(\mathrm{g}_{\phi}(x_k^a)), x_k^b\right),
\label{eq:recon-loss}
\end{equation}
where $\mathrm{g}_{\phi}$ is the encoder and $\mathrm{h}_{\theta}$ the decoder. The reconstruction loss $\mathrm{r(\cdot, \cdot)}$ is computed between the reconstruction of the input image $x_k^a$ and an image of the same scene under different environmental conditions $x_k^b$. In this work, we consider for the reconstruction loss the structural similarity index (SSIM) \cite{bergmann2018improving}: $\mathrm{r}(a,b)=1-\mathrm{SSIM}(a,b)$, but alternative image comparison functions can be considered as well. 

Our cost function formulation implies a partially impossible task to solve. The input image $x_k^a$ does not convey enough information to perfectly reconstruct the same scene under different environmental conditions $x_k^b$ in its entirety. While $x_k^a$ contains, usually, all the information of the objects in the scene, it does not contain any information about the illumination or environmental condition of $x_k^b$. However, both images are similar enough such that the encoder-decoder model can learn to focus on what is important, \ie the salient features (\eg people). Consequently, the only possibility for the neural network to minimize the loss is to focus on the objects in the scene which remain constant and neglect all the lightning and environment information, because the input images do not include information on how to handle it correctly. This implies that the neural network implicitly learns to focus the reconstruction on the people, objects and vehicle interior and to average out all the other information which changes between the similar scenes, \eg the illumination and environment. This can be observed in Fig. \ref{fig:sviro-recon-scene} where we compare the reconstruction of similar sceneries after training: all background information and lightning conditions has either been removed or replaced by constant values. The encoder learns to remove the illumination information. The decoder is light invariant and cannot produce different illuminations, since the information has already been removed in the latent space representation.

Our proposed method is not limited to having the same scenery under different illumination conditions. One could use different augmentation transformations on the same input image $x_k$ to form $x_k^a$ and $x_k^b$ and hence create the images on the fly. Alternatively, one could apply a \textit{reverse} denoising approach where only $x_k^b$ is augmented and $x_k^a$ is the clean input image. See Fig. S1 in the supplementary material for an example for both approaches.

\subsection{Triplet loss and nearest neighbour search}

\begin{figure}
	\begin{subfigure}{\linewidth}
  		\centering
  		\begin{overpic}[width=\linewidth]{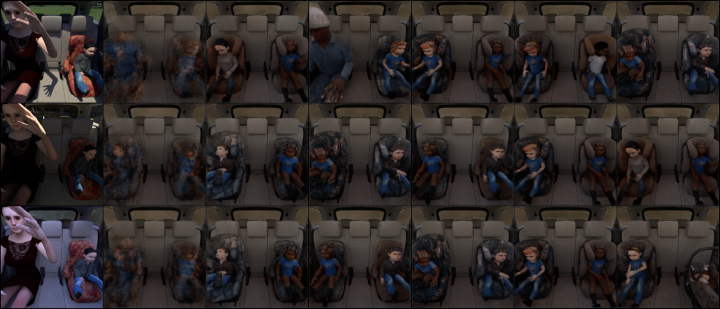}
	  		\put(3,39){\textcolor{white}{Input}}
	  		\put(16,39){\textcolor{white}{Recon}}
	  		\put(33,39){\textcolor{white}{1st}}
	  		\put(47,39){\textcolor{white}{2nd}}
	  		\put(61,39){\textcolor{white}{3rd}}
	  		\put(76,39){\textcolor{white}{4th}}
	  		\put(90,39){\textcolor{white}{5th}}
		\end{overpic}
  		\caption{Autoencoder}
	\end{subfigure}
	\vskip 0.2 \baselineskip
	\begin{subfigure}{\linewidth}
  		\centering
  		\includegraphics[width=\linewidth]{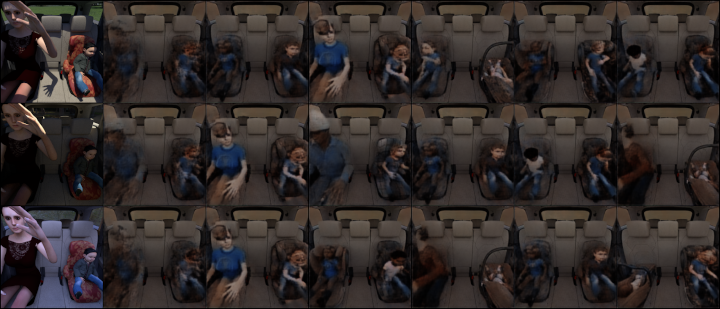}
  		\caption{Variational Autoencoder}
	\end{subfigure}
	\vskip 0.2 \baselineskip
	\begin{subfigure}{\linewidth}
  		\centering
  		\includegraphics[width=\linewidth]{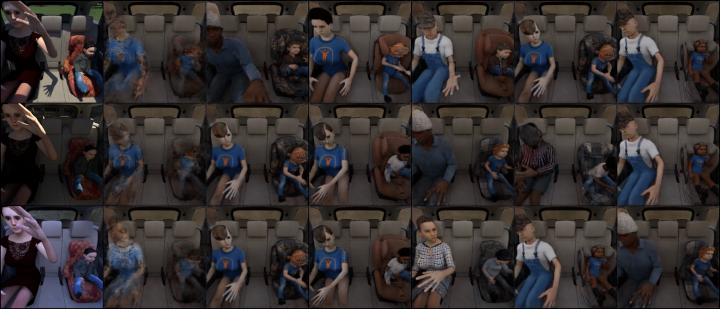}
  		\caption{Triplet Autoencoder}
	\end{subfigure}
	\caption{Comparison of the reconstruction of the 5 nearest neighbours (columns 3 to 7) for different encoder-decoder latent spaces (a), (b) and (c). The reconstruction (second column) of the test sample (first column) is also reported. The triplet regularization is by far the most reliable and consistent one across all 5 neighbours. Notice the class changes across neighbours for the AE and VAE models.}
	\label{fig:sviro-nn}
\end{figure}

While the aforementioned method works well on the training data, generalizing to unseen test images remains a challenging task if no additional precautions are taken. The illumination is still removed from test samples, but the reconstruction of the objects of interest can be less stable. As training data is limited, the encoder-decoder network is mostly used as a compression method instead of a generative model. Consequently, generalizing to unseen variations cannot trivially be achieved. Example of failures are plotted in Fig. \ref{fig:sviro-test-vs-nn} and Fig. \ref{fig:yale}: it can be observed, that the application on test images can cause blurry reconstructions. It turns out that the blurry reconstruction is in fact a blurry version of the reconstruction of its nearest neighbour in the latent space (or a combination of several nearest neighbours). An example of a comparison of the five nearest neighbours for several encoder-decoder models is shown in Fig. \ref{fig:sviro-nn}.

Consequently, instead of reconstructing the encoded test sample, it is more beneficial to reconstruct its nearest neighbour. However, applying nearest neighbour search in the latent space of a vanilla autoencoders (AE) or variational autoencoders (VAE) will not provide robust results. This is due to the fact that there is no guarantee that the learned latent space representation follows an $L^2$ metric \cite{arvanitidis2018latent}. As the nearest neighbour search is (usually) based on the $L^2$ norm, the latter will hence not always work reliably.

\begin{figure}
	\centering
	\includegraphics[width= \linewidth]{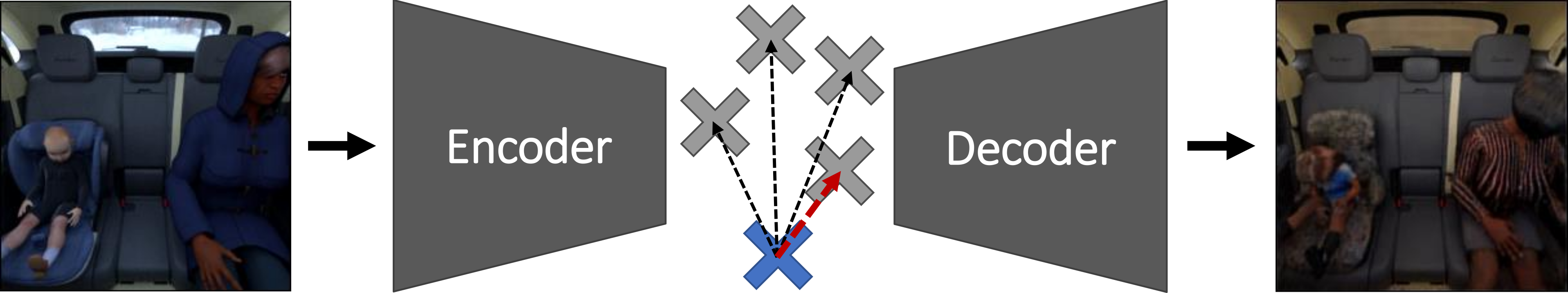}
	\caption{During inference, we choose the nearest neighbour (red arrow) of the latent space vector of the input image (blue cross) from all the training latent space vectors (gray crosses). This vector can be used to reconstruct a clean image or as classification prediction by using its label.}
	\label{fig:inference}
\end{figure}

To this end, we incorporated a triplet loss \cite{hoffer2015deep} in the latent space of the encoder-decoder model (TAE) instead. Using the same notations, the triplet loss can be defined as 
\begin{equation}
\begin{split}
	\mathcal{L}_T(\mathcal{X}_m; \phi) = &\sum_{k=0}^m \max \left(0, \norm{\mathrm{g}_{\phi}(x_k^{a,a})-\mathrm{g}_{\phi}(x_{k_1}^{a,p})}^2 \right. \\ 
	& \left . - \norm{\mathrm{g}_{\phi}(x_k^{a,a})-\mathrm{g}_{\phi}(x_{k_2}^{a,n})}^2 + \alpha \right),
\end{split}
\label{eq:triplet-loss}
\end{equation}
where $x_k^{a,a}$ is the anchor using scenery $k$, $x_{k_1}^{a,p}$ is the positive sample using a different scenery $k_1$ and $x_{k_2}^{a,n}$ is the negative sample using another scenery $k_2$. An illustration of the nearest neighbour inference is given in Fig. \ref{fig:inference} and for the triplet loss in Fig. S2. The triplet loss acts as a regularizer and due to its definition, it will also induce an $L^2$ norm in the latent space \cite{min2009deep, cosmo2020limp, arvanitidis2018latent}. This effect is highlighted in Fig. \ref{fig:sviro-nn}, where we compare the nearest neighbours of the AE, VAE and TAE. To take full advantage of the triplet selection, we also modified the reconstruction loss (\ref{eq:recon-loss}) such that it is computed for each of the triplet samples:
\begin{equation}
\begin{split}
	 &\mathcal{L}_R(\mathcal{X}_m; \theta, \phi) = \sum_{k=0}^{m} \mathrm{r}\left(\mathrm{h}_{\theta}(\mathrm{g}_{\phi}(x_k^{a,a})), x_k^{b,a}\right) \\
	 & + \mathrm{r}\left(\mathrm{h}_{\theta}(\mathrm{g}_{\phi}(x_{k_1}^{a,p})), x_{k_1}^{b,p}\right) + \mathrm{r}\left(\mathrm{h}_{\theta}(\mathrm{g}_{\phi}(x_{k_2}^{a,n})), x_{k_2}^{b,n}\right),
\end{split}
\label{eq:recon-triplet-loss}
\end{equation}
where we take for each input image $x_{\cdot}^{a,\cdot}$ a different random output image $x_{\cdot}^{b,\cdot}$. Consequently, the total loss is defined as 
\begin{equation}
	 \mathcal{L}(\mathcal{X}_m; \theta, \phi) = \mathcal{L}_R(\mathcal{X}_m; \theta, \phi) + \mathcal{L}_T(\mathcal{X}_m; \phi).
\label{eq:total-loss}
\end{equation}

\section{Experiments}
We will present an analysis of the aforementioned properties, problems and improvements on the SVIRO-Illumination dataset to highlight the benefit of our design choices. We will present results on two additional publicly available datasets to show the applicability of our proposed cost function to other problem formulations as well. 

\subsection{Training details}

We center-cropped the images to the smallest image dimension and then resized it to a size of 224x224. We used a batch size of 16, trained our models for 1000 epochs and did not perform any data augmentation. We used the Adam optimizer and a learning rate of 0.0001. Image similarity between target image and reconstruction was computed using SSIM \cite{bergmann2018improving}. We used a latent space of dimension 16. The model architecture is detailed in Table S1 in the supplementary material: it uses the VGG-11 architecture \cite{simonyan2014very} for the encoder part and reverses the layers together with nearest neighbour up-sampling for the decoder part. However, our proposed cost function is not limited to the model's architecture choice. We used PyTorch 1.6, torchvision 0.7 and pytorch-msssim 2.0 \cite{Gongfan2019} for all our experiments. 

\subsection{SVIRO-Illumination}
\begin{figure}
	\centering
	\includegraphics[width=0.24\linewidth]{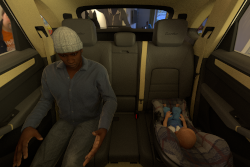}
	\includegraphics[width=0.24\linewidth]{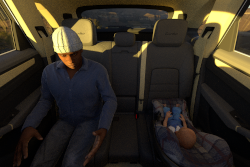}
	\includegraphics[width=0.24\linewidth]{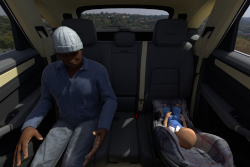}
	\includegraphics[width=0.24\linewidth]{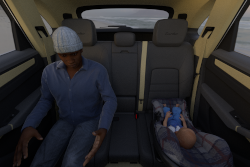}
	\vskip 0.1 \baselineskip
	\includegraphics[width=0.24\linewidth]{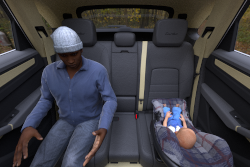}
	\includegraphics[width=0.24\linewidth]{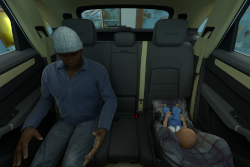}
	\includegraphics[width=0.24\linewidth]{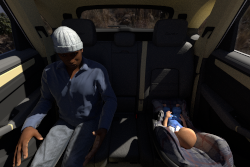}
	\includegraphics[width=0.24\linewidth]{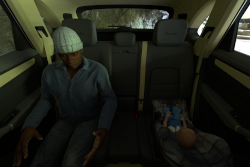}
	\caption{Example scenery from SVIRO-Illumination. The same scenery under eight (out of ten) different illumination and external environments. Left seat: adult passenger, middle seat:empty and right seat: infant seat with a baby.}
	\label{fig:example-sviro}
\end{figure}
Based on the recently released SVIRO dataset \cite{DiasDaCruz2020SVIRO}, we created additional images for three new vehicle interiors. For each vehicle, we randomly generated 250 training and 250 test scenes where each scenery was rendered under 10 different illumination and environmental conditions. We created two versions: one containing only people and a second one including additionally occupied child and infant seats. We used 10 different exterior environments (HDR images rotated randomly around the vehicles), 14 (or 8) human models, 6 (or 4) children and 3 babies respectively for the training and test split. The four infant and two child seats have the same geometry for each split, but they use different textures. Consequently, the models need to generalize to new illumination conditions, humans and textures. There are four possible classes for each seat position (empty, infant seat, child seat and adult) leading to a total of $4^3=64$ classes for the whole image. Examples are shown in Fig. \ref{fig:example-sviro} and Fig. S3-S5 in the supplementary material.

\subsubsection{Reconstruction results}
For the triplet loss sampling, we chose the positive sample to be of the same class as the anchor image (but from a different scenery) and the negative sample to differ only on one seat (\ie change only the class on a single seat w.r.t. the anchor image). Images of three empty seats do no contain any information which could mislead the network, so to make it more challenging, we did not use them as negative samples. 

\begin{figure}
	\begin{overpic}[width=\linewidth]{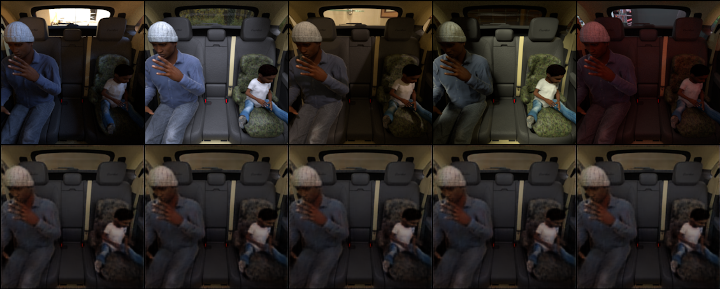}
		\put(1,25){\rotatebox{90}{\textcolor{white}{Input}}}
		\put(1,4){\rotatebox{90}{\textcolor{white}{Recon}}}
	\end{overpic}
	\caption{The encoder-decoder model transforms the input images of a same scenery (first row) into a cleaned version (second row) by removing all illumination and environment information (see the background through the window)}
	\label{fig:sviro-recon-scene}
\end{figure}

\begin{figure}
	\begin{overpic}[width=\linewidth]{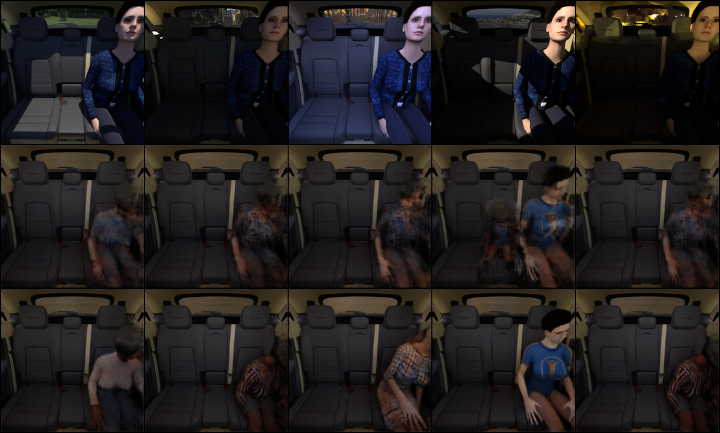}
		\put(1,46){\rotatebox{90}{\textcolor{white}{Input}}}
		\put(1,25){\rotatebox{90}{\textcolor{white}{Recon}}}
		\put(1,1){\rotatebox{90}{\textcolor{white}{NN-Recon}}}
	\end{overpic}
	\caption{The test image (first row) cannot be reconstructed perfectly (second row). However, choosing the nearest neighbour in the latent space and reconstructing the latter leads to a class preserving reconstruction (third row).}
	\label{fig:sviro-test-vs-nn}
\end{figure}

\begin{figure}
	\centering
	\begin{subfigure}{0.23\linewidth}
 		\centering
  		\includegraphics[width=\linewidth]{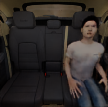}
  		\caption*{\footnotesize{First recon}}
	\end{subfigure}\hfill
	\begin{subfigure}{0.23\linewidth}
  		\centering
  		\includegraphics[width=\linewidth]{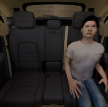}
  		\caption*{\footnotesize{Max recon}}
	\end{subfigure}\hfill
	\begin{subfigure}{0.23\linewidth}
  		\centering
  		\includegraphics[width=\linewidth]{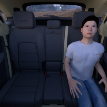}
  		\caption*{\footnotesize{Closest input}}
	\end{subfigure}\hfill
	\begin{subfigure}{0.23\linewidth}
  		\centering
  		\includegraphics[width=\linewidth]{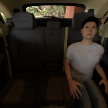}
  		\caption*{\footnotesize{Furthest input}}
	\end{subfigure}
	\caption{The reconstruction of the first scenery input image (first recon) is compared against the furthest reconstruction of all scenery input images (max recon). First recon is also used to determine the closest and furthest scenery input images. The encoder-decoder model does not learn to focus the reconstruction to a training sample. }
	\label{fig:sviro-extrema}
\end{figure}

After training, the encoder-decoder model learned to remove all the illumination and environmental information from the training images. See Fig. \ref{fig:sviro-recon-scene} for an example on how images from the same scenery, but under different illumination, are transformed. Sometimes, test samples are not reconstructed reliably. However, due to the triplet constraint and nearest neighbour search, we can preserve the correct classes and reconstruct a clean image: see Fig. \ref{fig:sviro-test-vs-nn} for an example. The reconstruction of the test image latent vector produces a blurry person, which is usually a combination of several nearest neighbours. The reliability of the class preservations is investigated in Section \ref{sec:classfication} based on the classification accuracy.  We want to emphasize that the model is not learning to focus the reconstruction to a single training image for each scenery. In Fig. \ref{fig:sviro-extrema} we searched for the closest and furthest (w.r.t. SSIM) input images of the selected scenery w.r.t to the reconstruction of the first input image. Moreover, we selected the reconstruction of all input images which is furthest away from the first one to get an idea about the variability of the reconstructions inside a single scenery. While the reconstructions are stable for all images of a scenery, it can be observed that the reconstructions are far from all training images. Hence, the model did not learn to focus the reconstruction to a single training sample, but instead learned to remove all the unimportant information from the input image. The shape and features of the salient objects are preserved as long as their position remains constant in each image, \eg see Fig. \ref{fig:webcam} for vehicles being removed if not contained in each image. The texture of the salient objects is uniformly lit and smoothed out.

\subsubsection{AE vs. VAE vs. TAE}

\begin{figure*}
	\centering
	\begin{subfigure}{0.30\linewidth}
 		\centering
  		\includegraphics[width=\linewidth]{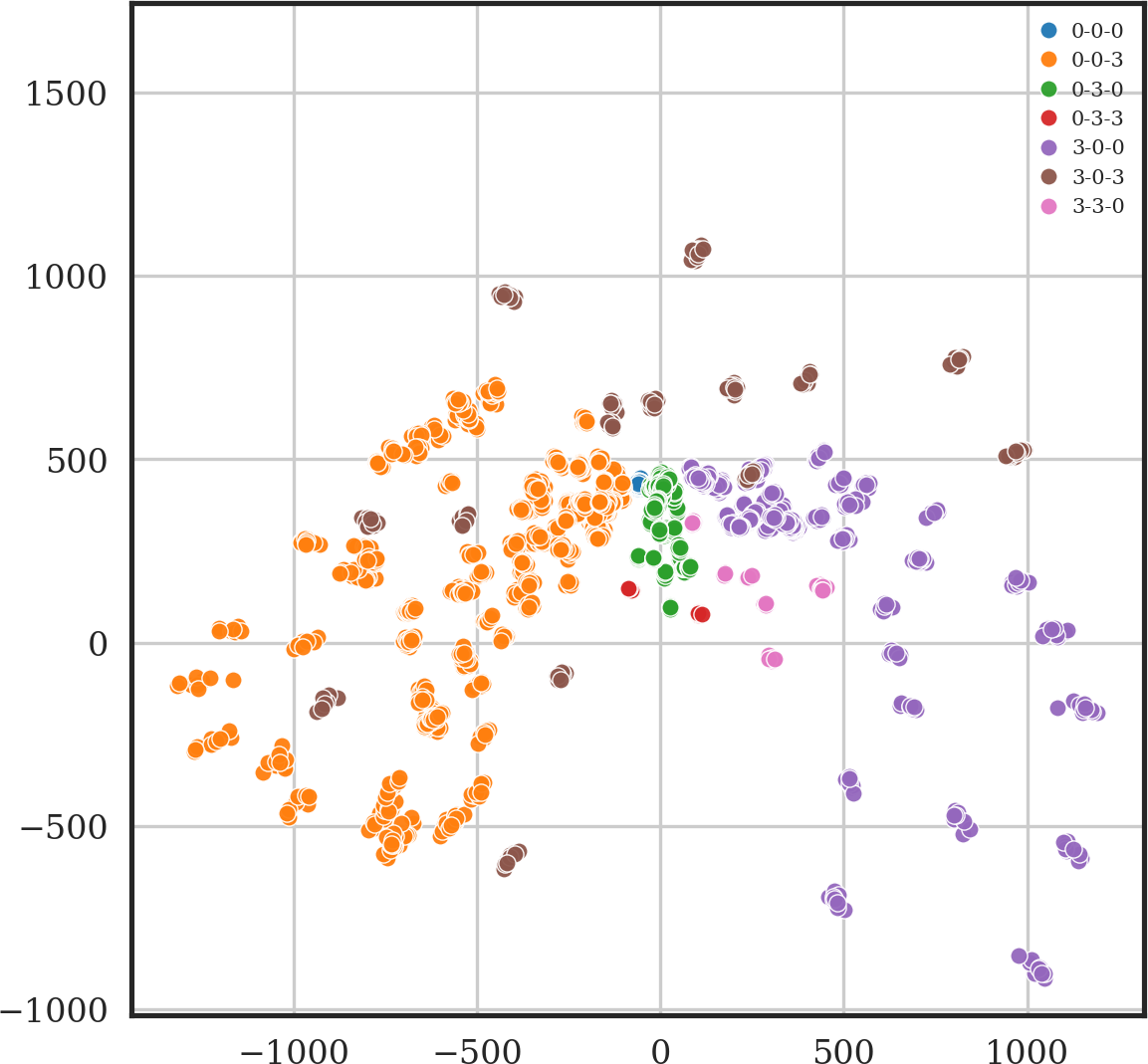}
  		\caption{Autoencoder}
	\end{subfigure}\hfill
	\begin{subfigure}{0.30\linewidth}
  		\centering
  		\includegraphics[width=\linewidth]{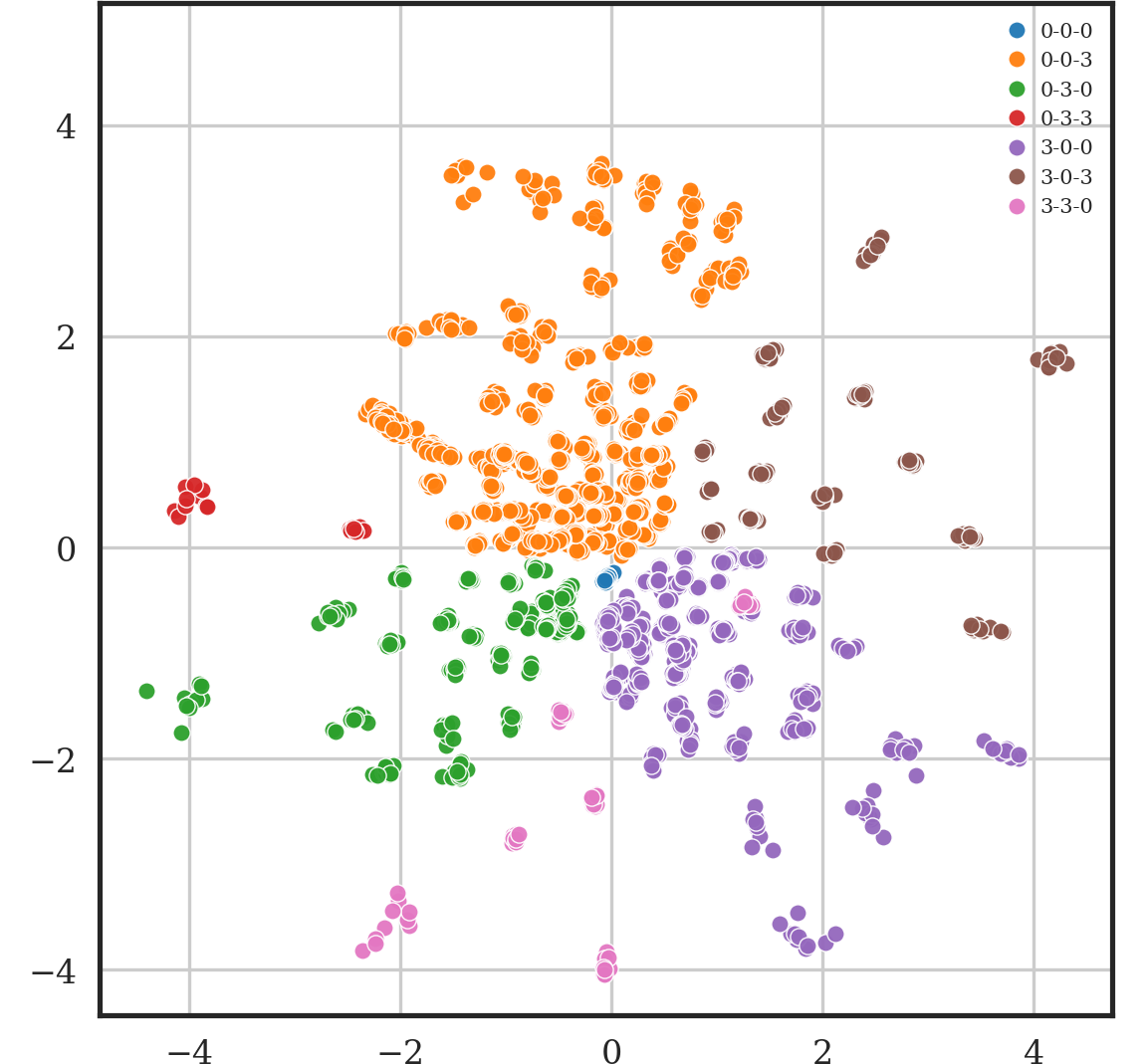}
  		\caption{Variational Autoencoder}
	\end{subfigure}\hfill
	\begin{subfigure}{0.30\linewidth}
  		\centering
  		\includegraphics[width=\linewidth]{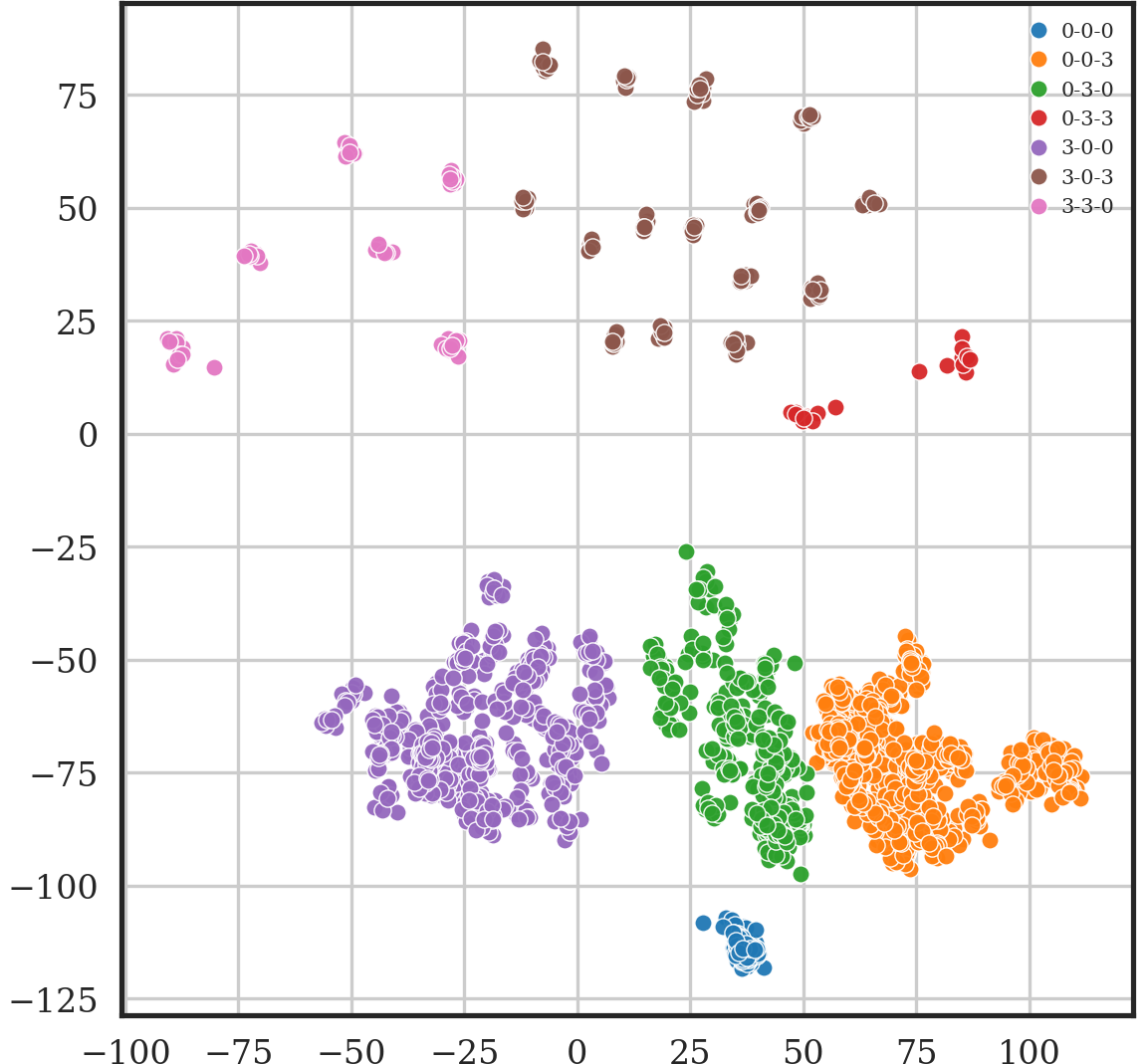}
  		\caption{Triplet Autoencoder}
	\end{subfigure}
	\caption{Comparison of training data latent space distributions for different regularizers in the latent space of encoder-decoder models. Different colors represent different classes. For each seat position, we either have 0 (empty) or 3 (adult) such that an image is a composition of three labels, e.g. for 3-0-3 an adult is sitting left and right. Some classes are under-represented and some samples are clustered close together: those are identical sceneries under different lightning conditions.}
	\label{fig:latent}
\end{figure*}

For visualization purposes, we trained a vanilla autoencoder (AE), variational autoencoder (VAE) and triplet autoencoder (TAE) on the SVIRO-Illumination dataset with people and empty seats only. For simplicity of visualization, we chose a latent space dimension of 2 for the model definition. After training, we computed the latent space representation for all training samples and plotted the resulting distributions in Fig. \ref{fig:latent}. The triplet based encoder-decoder model separates and clusters the classes best. Some small clusters are due to under-represented classes, for which the model clusters images from the same scenery under different illuminations together. The AE uses a large range of possible values in the latent space and both the AE and VAE contain wrong classes inside other clusters. The test distribution is plotted in Fig. S6 in the supplementary material and highlights the additional benefit of the TAE for potential outlier detection. Moreover, we show in Fig. S7 and Fig. S8 that a 2-dimensional principal component analysis and T-SNE projection of a 16-dimensional latent space provides even further benefits when a TAE is used. The same models were trained with a latent space dimension of 16 including occupied child and infant seats. The classification results obtained by nearest neighbour search are compared against several other models in Section \ref{sec:classfication}. The TAE outperforms the other encoder-decoder models w.r.t. accuracy. 

We needed to adjust the weight in the loss for the KL divergence (regularizer w.r.t. Gaussian prior) to $\beta=0.001$ for training the VAE and prevent mode collapses. This is due to the background of the vehicle interior which is dominant in all training samples and remains similar.  

It is important to note that the comparison between AE, VAE and TAE is not entirely fair, because the latter implicitly uses labels during the positive and negative sample selection. Nevertheless, for the problem formulations at hand, it is beneficial to collect the classification labels  considering the additional advantage of the induced $L^2$ norm in the latent space and improved classification accuracy.

\subsubsection{Classification results}
\label{sec:classfication}
We further compared the classification accuracy of our proposed method together with the nearest neighbour search against vanilla classification models when the same training data is being used. This way, we can quantitatively estimate the reliability of our proposed method against commonly used models. To this end, we trained baseline classification models (ResNet-50 \cite{he2016deep}, VGG-11 \cite{simonyan2014very} and MobileNet V2 \cite{sandler2018mobilenetv2}) as pre-defined in torchvision on SVIRO-Illumination. For each epoch, we randomly selected one $x_k^j \in \mathcal{X}$ for each scenery $x_k$. The classification models were either trained for 1000 epochs or we performed early stopping with a 80:20 split on the training data. We further fine-tuned pre-trained models for 1000 epochs. The triplet based autoencoder model is being trained exactly as before. During inference, we take the label of the nearest training sample as the classification prediction. The random seeds of all libraries were fixed for all experiments and cuDNN was used in deterministic mode. Each setup was repeated 5 times with 5 different (but the same ones across all setups) seeds. Moreover, the experiments are repeated for all three vehicle interiors. The mean classification accuracy over all 5 runs together with the variance is reported in Table \ref{table:compare_classif}. Our proposed method significantly outperforms vanilla classification models trained from scratch and the models' performances undergo a much smaller variance. Moreover, our proposed method outperforms fine-tuned pre-trained classification models, despite the advantage of the pre-training of these models. Additionally, we trained the encoder-decoder models using the vanilla reconstruction error between input and reconstruction, but using the nearest neighbour search as a prediction. Again, including our proposed reconstruction loss improves the models' performance significantly.

\begin{table}
	\caption{Mean accuracy and variance over 5 repeated training runs on each of the three vehicle interiors. F = fine-tuned pre-trained model, ES=early stopping with 80:20 split, NS=no early stopping and V=vanilla reconstruction loss. Our proposed reconstruction loss improves the encoder-decoder vanilla version and with the nearest neighbour search outperforms all other models significantly.}
	\begin{center}
		\def\arraystretch{1.3}
		\begin{tabular}{|c|ccc|}
			\cline{2-4}
			\multicolumn{1}{c|}{} & \multicolumn{3}{c|}{Vehicle}\\
			\hline
			Model & Cayenne & Kodiaq & Kona\\
			\hline
			\hline
			MobileNet-ES & 62.9 $\pm$ 3.1 & 71.8 $\pm$ 4.3 & 73.0 $\pm$ 0.8 \\
			\hline
			VGG11-ES & 64.4 $\pm$ 35 & 74.0 $\pm$ 19 & 75.5 $\pm$ 5.7 \\
			\hline
			ResNet50-ES & 72.3 $\pm$ 3.7 & 77.9 $\pm$ 35 & 76.6 $\pm$ 9.9 \\
			\hline
			\hline
			MobileNet-NS & 72.7 $\pm$ 3.8 & 77.0 $\pm$ 4.1 & 77.4 $\pm$ 2.2 \\
			\hline
			VGG11-NS & 74.1 $\pm$ 5.8 & 71.2 $\pm$ 14 & 78.4 $\pm$ 2.6 \\
			\hline
			ResNet50-NS & 76.2 $\pm$ 18 & 83.1 $\pm$ 1.1 & 82.0 $\pm$ 3.2 \\
			\hline
			\hline
			MobileNet-F & 85.8 $\pm$ 2.0 & 90.6 $\pm$ 1.2 & 88.6 $\pm$ 0.6 \\
			\hline
			VGG11-F & 90.5 $\pm$ 2.0 & 90.3 $\pm$ 1.2 & 89.2 $\pm$ 0.9 \\
			\hline
			ResNet50-F & 87.9 $\pm$ 2.0 & 89.7 $\pm$ 6.1 & 88.5 $\pm$ 1.0 \\
			\hline
			\hline
			AE-V & 74.1 $\pm$ 0.7 & 80.1 $\pm$ 1.8 & 73.3 $\pm$ 0.9 \\
			\hline
			VAE-V & 73.4 $\pm$ 1.3 & 79.5 $\pm$ 0.6 & 73.0 $\pm$ 0.9 \\
			\hline
			TAE-V & \underline{90.8} $\pm$ 0.3 & \underline{91.7} $\pm$ 0.2 & \underline{89.9} $\pm$ 0.6 \\
			\hline
			\hline
			AE (ours) & 86.8 $\pm$ 0.3 & 86.7 $\pm$ 1.5 & 86.7 $\pm$ 0.9 \\			
			\hline
			VAE (ours) & 81.4 $\pm$ 0.5 & 86.6 $\pm$ 0.9 & 85.9 $\pm$ 0.8 \\
			\hline
			TAE (ours) & \underline{\textbf{92.4}} $\pm$ 1.5 & \underline{\textbf{93.5}} $\pm$ 0.9 & \underline{\textbf{93.0}} $\pm$ 0.3 \\
			\hline
		\end{tabular}
	\end{center}
	\label{table:compare_classif}
\end{table}

\subsection{Extended Yale Face Database B}

\begin{figure}
	\centering
	
	\begin{overpic}[width=0.19\linewidth]{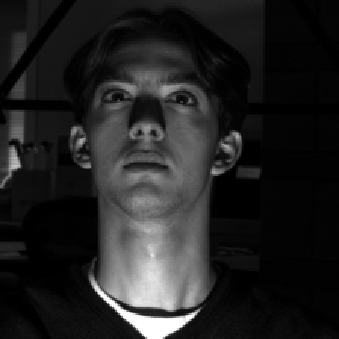}
		\put(7,28){\rotatebox{90}{\textcolor{white}{Input}}}
	\end{overpic}
	\includegraphics[width=0.19\linewidth]{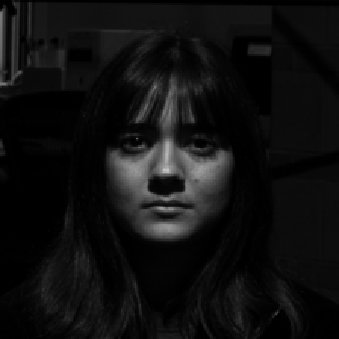}
	\includegraphics[width=0.19\linewidth]{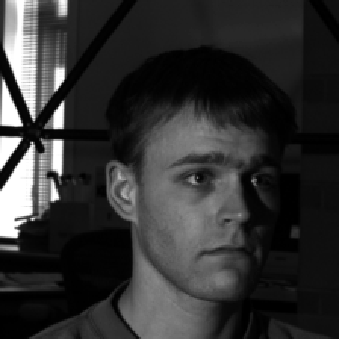}
	\includegraphics[width=0.19\linewidth]{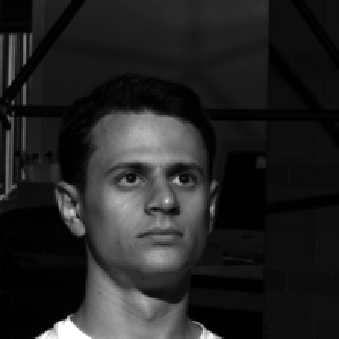}
	\includegraphics[width=0.19\linewidth]{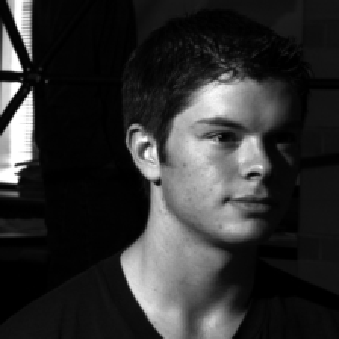}
	
	\vskip 0.1 \baselineskip	
	
	\begin{overpic}[width=0.19\linewidth]{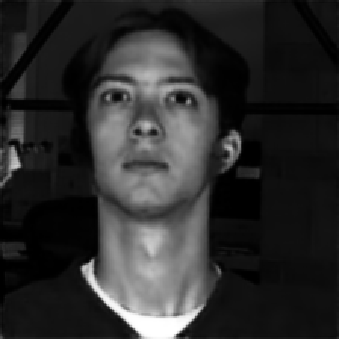}
		\put(7,22){\rotatebox{90}{\textcolor{white}{Recon}}}
	\end{overpic}
	\includegraphics[width=0.19\linewidth]{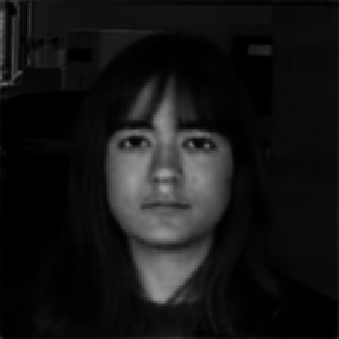}
	\includegraphics[width=0.19\linewidth]{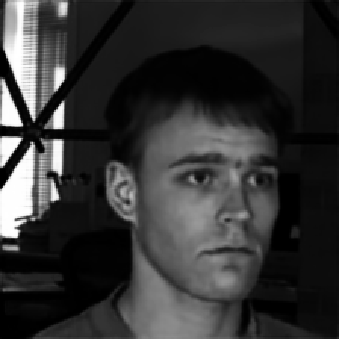}
	\includegraphics[width=0.19\linewidth]{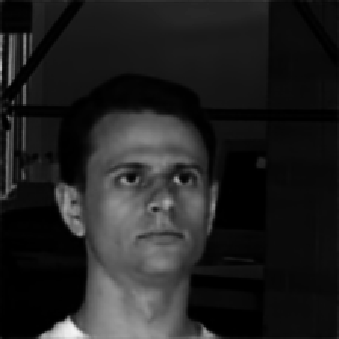}
	\includegraphics[width=0.19\linewidth]{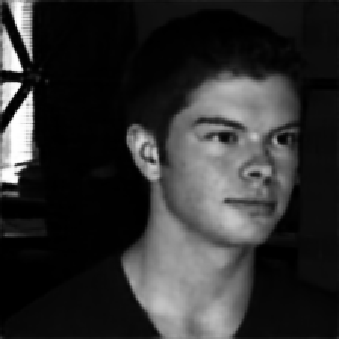}
	
	\vskip 0.1 \baselineskip	
	
	\begin{overpic}[width=0.19\linewidth]{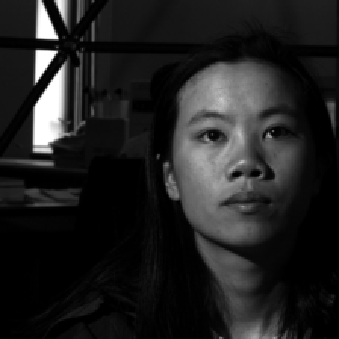}
		\put(7,28){\rotatebox{90}{\textcolor{white}{Input}}}
	\end{overpic}
	\includegraphics[width=0.19\linewidth]{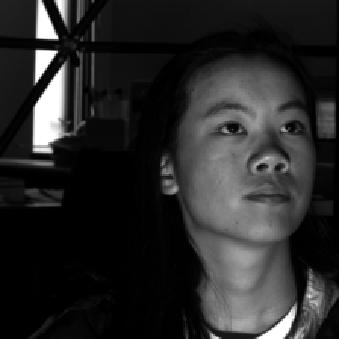}
	\includegraphics[width=0.19\linewidth]{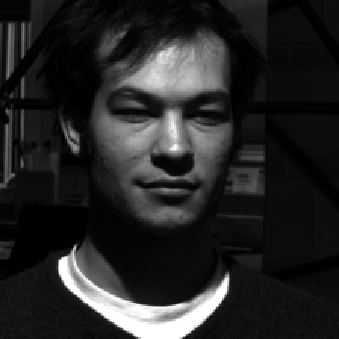}
	\includegraphics[width=0.19\linewidth]{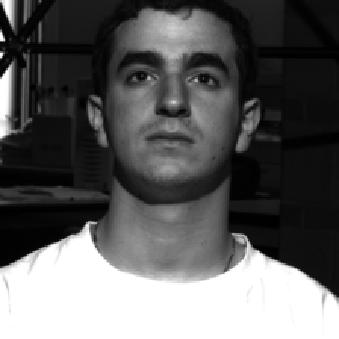}
	\includegraphics[width=0.19\linewidth]{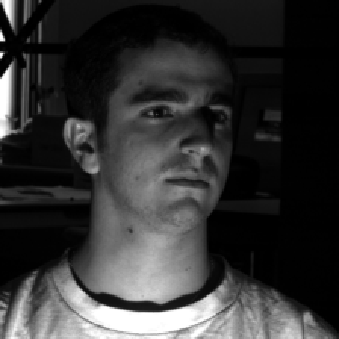}
	
	\vskip 0.1 \baselineskip
	
	\begin{overpic}[width=0.19\linewidth]{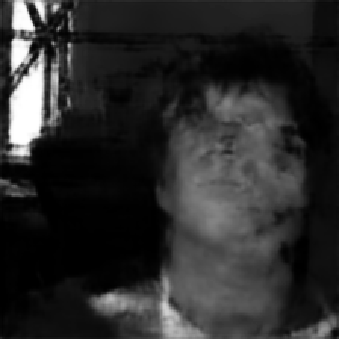}
		\put(7,22){\rotatebox{90}{\textcolor{white}{Recon}}}
	\end{overpic}
	\includegraphics[width=0.19\linewidth]{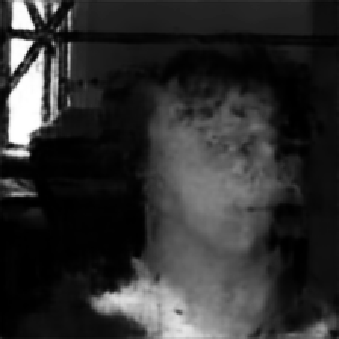}
	\includegraphics[width=0.19\linewidth]{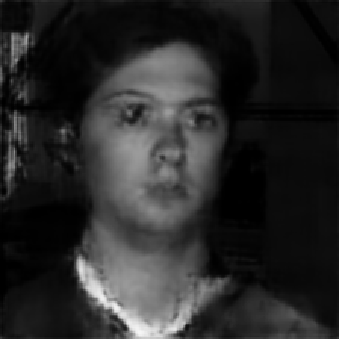}
	\includegraphics[width=0.19\linewidth]{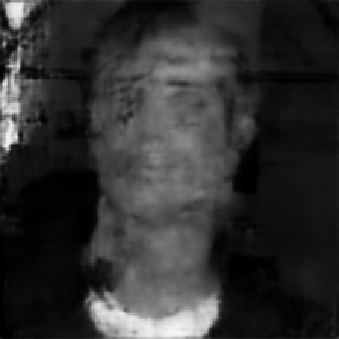}
	\includegraphics[width=0.19\linewidth]{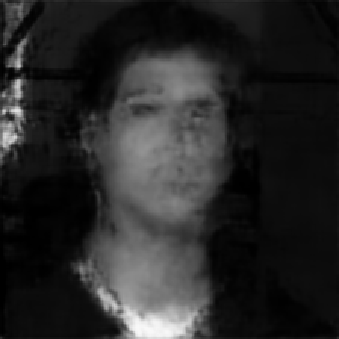}
	
	\vskip 0.1 \baselineskip	
	
	\begin{overpic}[width=0.19\linewidth]{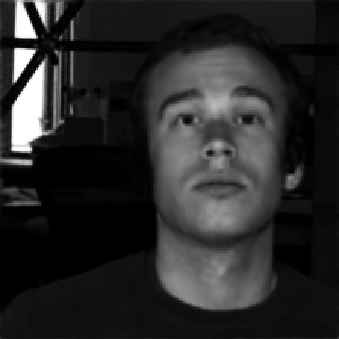}
		\put(7,1){\rotatebox{90}{\textcolor{white}{NN-Recon}}}
	\end{overpic}
	\includegraphics[width=0.19\linewidth]{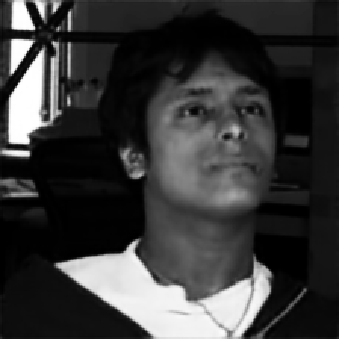}
	\includegraphics[width=0.19\linewidth]{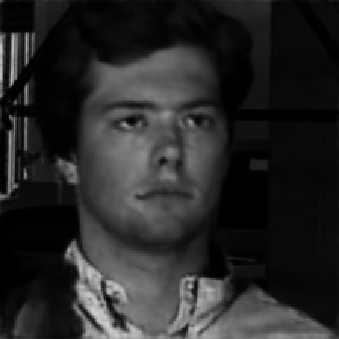}
	\includegraphics[width=0.19\linewidth]{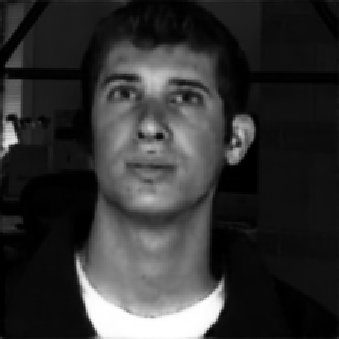}
	\includegraphics[width=0.19\linewidth]{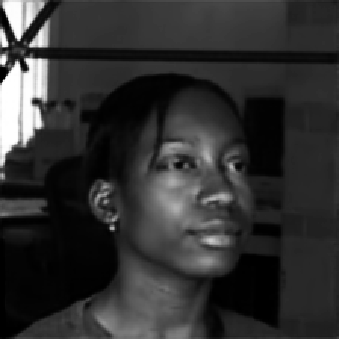}
	
	\caption{Examples for Extended Yale Face Database B. All the illumination information is removed from the training samples (first row) to form the reconstruction (second row). The test samples (third row) cannot always be reconstructed reliably (fourth row). However, by reconstructing the nearest neighbour (fifth row) the head pose and position of the head can be preserved and the illumination removed.}
	\label{fig:yale}
\end{figure}

The Extended Yale Face Database B \cite{GeBeKr01} contains images of 28 human subjects under 9 poses. For each pose and human subject, the same image is recorded under 64 illumination conditions. We considered the full-size image version instead of the cropped one and used 25 human subjects for training and 3 for the testing. We removed some of the extreme dark (no face visible) illumination conditions. Example images from the dataset are plotted in Fig. \ref{fig:yale}.

For the triplet sampling we chose as a positive sample an image with the same head pose and for the negative sample an image with a different head pose. We report qualitative results of a trained model in Fig. \ref{fig:yale}. The model is able to remove lightning and shadows from the training images, but the vanilla reconstruction on test samples can be blurry. We are not using the center cropped variant of the dataset, which makes the task more complicated, because the head is not necessarily at the same position for different human subjects. Nevertheless, the model is able to provide a nearest neighbour with a similar head pose and head position.

\subsection{Webcam Clip Art}
\begin{figure}
	\centering
	
	\begin{overpic}[width=0.19\linewidth]{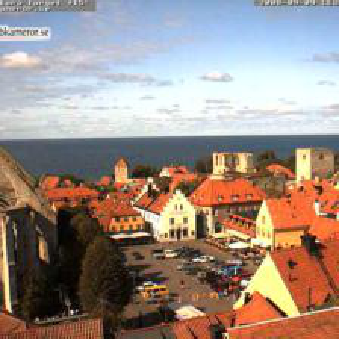}
		\put(7,28){\rotatebox{90}{\textcolor{white}{Input}}}
	\end{overpic}
	\hfill
	\includegraphics[width=0.19\linewidth]{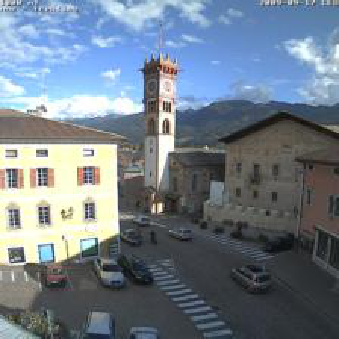}
	\hfill
	\includegraphics[width=0.19\linewidth]{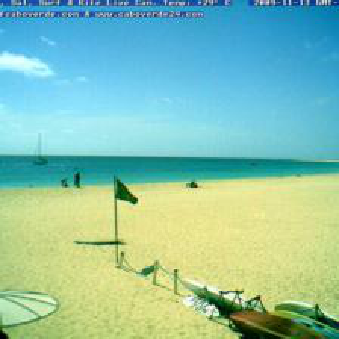}
	\hfill
	\includegraphics[width=0.19\linewidth]{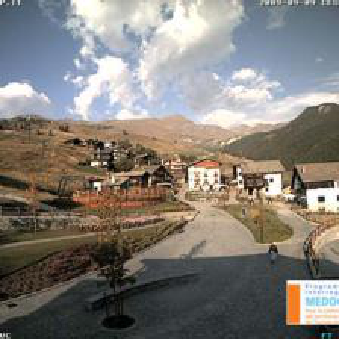}
	\hfill
	\includegraphics[width=0.19\linewidth]{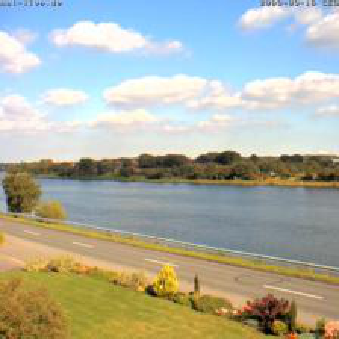}
	
	\vskip 0.1 \baselineskip
	
	\begin{overpic}[width=0.192\linewidth]{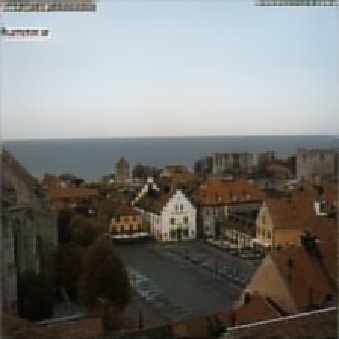}
		\put(7,22){\rotatebox{90}{\textcolor{white}{Recon}}}
	\end{overpic}
	\hfill
	\includegraphics[width=0.19\linewidth]{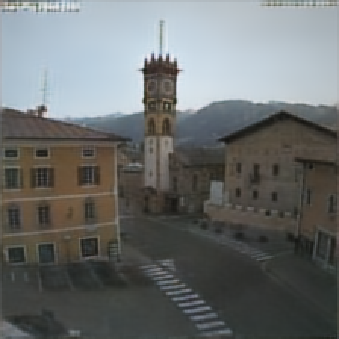}
	\hfill
	\includegraphics[width=0.19\linewidth]{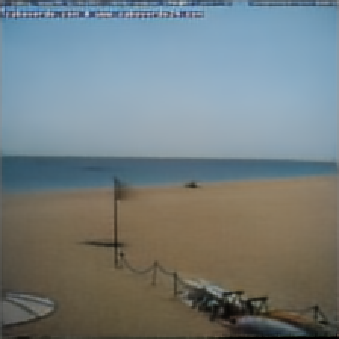}
	\hfill
	\includegraphics[width=0.19\linewidth]{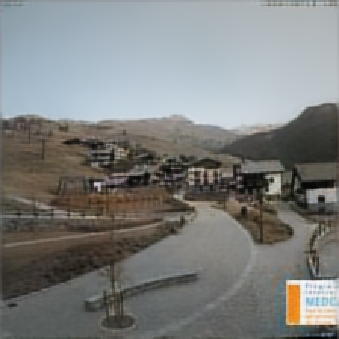}
	\hfill
	\includegraphics[width=0.19\linewidth]{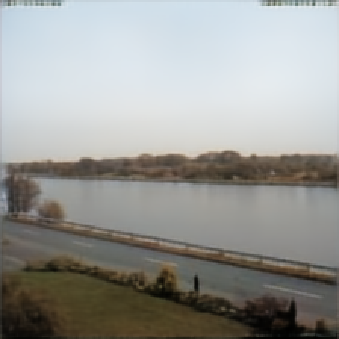}
	
	\caption{Examples for the Webcam Clip Art dataset. The encoder-decoder model removes the environmental features from the images (first row) to form the output images (second row). Vehicles and people are removed from the scenery and skies, rivers and beaches are smoothed out.}
	\label{fig:webcam}
\end{figure}

The Webcam Clip Art \cite{lalonde-siggraph-asia-09} dataset consists of images from 54 webcams from places all over the world. The images are recorded continuously such that a same scenery is available for different day times, seasons and weather conditions. For each of the 54 regions, we selected randomly 100 sceneries. Example images are provided in Fig. \ref{fig:webcam}. 

For the triplet sampling, we chose as positive sample an image from the same location and for the negative sample an image from a different location. Each landscape and building arrangement undergoes unique shadow, illumination and reflection properties. The generalization to unknown places under unknown illumination conditions is thus too demanding to be deduced from a single input image. Hence, we do not provide a test evaluation and report results on training samples only in Fig. \ref{fig:webcam}. The model removes the illumination variations and shadows from the images. Moreover, rivers, oceans and skies as well as beaches are smoothed out. Most of the people and cars are removed and replaced by the actual background of the scenery.

\section{Limitations}

Our proposed method works well on the training data, which can be sufficient for some applications, \eg when a fixed dataset is available on which some post-processing needs to be done only. Since the generalization to test images can be achieved by a nearest neighbour search, the latter will only be useful for a subset of machine learning tasks. Our method preserves the classes for a given problem formulation, which will be fine for classification and object detection. Although our method preserves even head poses (e.g. Fig. \ref{fig:yale}) when it is dominantly present in the training images, our approach will likely not preserve complex human poses (e.g. Fig. \ref{fig:sviro-test-vs-nn}) or detailed facial landmarks, because the body poses and key features are not necessarily preserved by the nearest neighbour search. Future work should try to incorporate constraints such that the poses and landmarks of test samples are preserved as well.

In practise, it will be challenging to record identical sceneries under different lightning conditions. However, as the Extended Yale Face Database B \cite{GeBeKr01} and Webcam Clip Art \cite{lalonde-siggraph-asia-09} dataset have shown, it is also feasible. Since we have highlighted the benefit of the acquisition of said datasets, the investment of recording under similar conditions in practise can be worth for some applications. We believe that future work will develop possibilities to facilitate the data acquisition process. Moreover, the possibility to incorporate images taken for the same scene, but in less perfect conditions, should be explored (e.g. Fig. S1).

\section{Conclusion}
Our results show the benefit of recording identical sceneries under different lightning and environmental conditions such that unwanted features can be remove by a partially impossible reconstruction loss function: without the need for a ground truth target image. Our method works well for classification and post-processing tasks due to an enhanced nearest neighbour search induced by a triplet loss regularization in the latent space of an encoder-decoder network. We demonstrated the universal applicability of our proposed method, as long as the correct data (i.e. same scenery under different conditions) is available, on three different tasks and datasets. Moreover, our proposed method improves classification accuracy significantly compared to standard encoder-decoder and classification models, even when the latter was a fine-tuned pre-trained model.

\section*{Acknowledgement}
The first author is supported by the Luxembourg National Research Fund (FNR) under grant number 13043281. The second author is supported by VIDETE (grant number 01IW18002). This work was partially funded by the Luxembourg Ministry of the Economy (CVN 18/18/RED).

{\small
\bibliographystyle{ieee_fullname}
\bibliography{bib}

\begin{thebibliography}{10}\itemsep=-1pt

\bibitem{alshammari2018impact}
Naif Alshammari, Samet Akcay, and Toby~P Breckon.
\newblock On the impact of illumination-invariant image pre-transformation for
  contemporary automotive semantic scene understanding.
\newblock In {\em Intelligent Vehicles Symposium (IV)}, 2018.

\bibitem{antelmi2019sparse}
Luigi Antelmi, Nicholas Ayache, Philippe Robert, and Marco Lorenzi.
\newblock Sparse multi-channel variational autoencoder for the joint analysis
  of heterogeneous data.
\newblock In {\em International Conference on Machine Learning (PMLR)}, 2019.

\bibitem{arvanitidis2018latent}
Georgios Arvanitidis, Lars~Kai Hansen, and S{\o}ren Hauberg.
\newblock Latent space oddity: on the curvature of deep generative models.
\newblock In {\em International Conference on Learning Representations (ICLR)},
  2018.

\bibitem{badino2011visual}
Hern{\'a}n Badino, D Huber, and Takeo Kanade.
\newblock Visual topometric localization.
\newblock In {\em Intelligent Vehicles Symposium (IV)}, 2011.

\bibitem{bergmann2018improving}
Paul Bergmann, Sindy L{\"o}we, Michael Fauser, David Sattlegger, and Carsten
  Steger.
\newblock Improving unsupervised defect segmentation by applying structural
  similarity to autoencoders.
\newblock {\em arXiv preprint arXiv:1807.02011}, 2018.

\bibitem{cosmo2020limp}
Luca Cosmo, Antonio Norelli, Oshri Halimi, Ron Kimmel, and Emanuele Rodol{\`a}.
\newblock Limp: Learning latent shape representations with metric preservation
  priors.
\newblock {\em arXiv preprint arXiv:2003.12283}, 2020.

\bibitem{DiasDaCruz2020SVIRO}
Steve {Dias Da Cruz}, Oliver Wasenm\"uller, Hans-Peter Beise, Thomas Stifter,
  and Didier Stricker.
\newblock Sviro: Synthetic vehicle interior rear seat occupancy dataset and
  benchmark.
\newblock In {\em IEEE Winter Conference on Applications of Computer Vision
  (WACV)}, 2020.

\bibitem{GeBeKr01}
A.S. Georghiades, P.N. Belhumeur, and D.J. Kriegman.
\newblock From few to many: Illumination cone models for face recognition under
  variable lighting and pose.
\newblock {\em IEEE Transactions on Pattern Analysis and Machine Intelligence
  (PAMI)}, 2001.

\bibitem{Glover2010ICRA}
Arren Glover, Will Maddern, Michael Milford, and Gordon Wyeth.
\newblock {FAB-MAP + RatSLAM: Appearance-based SLAM for Multiple Times of Day}.
\newblock In {\em Conference on Robotics and Automation (ICRA)}, 2010.

\bibitem{glover2010fab}
Arren~J Glover, William~P Maddern, Michael~J Milford, and Gordon~F Wyeth.
\newblock Fab-map+ ratslam: Appearance-based slam for multiple times of day.
\newblock In {\em International Conference on Robotics and Automation (ICRA)},
  2010.

\bibitem{Gongfan2019}
Fang Gongfan.
\newblock Pytorch ms-ssim.
\newblock \url{https://github.com/VainF/pytorch-msssim}, 2019.

\bibitem{he2016deep}
Kaiming He, Xiangyu Zhang, Shaoqing Ren, and Jian Sun.
\newblock Deep residual learning for image recognition.
\newblock In {\em Conference on Computer Vision and Pattern Recognition
  (CVPR)}, 2016.

\bibitem{hoffer2015deep}
Elad Hoffer and Nir Ailon.
\newblock Deep metric learning using triplet network.
\newblock In {\em International Workshop on Similarity-Based Pattern
  Recognition (SIMBAD)}, 2015.

\bibitem{karras2018progressive}
Tero Karras, Timo Aila, Samuli Laine, and Jaakko Lehtinen.
\newblock Progressive growing of gans for improved quality, stability, and
  variation.
\newblock In {\em International Conference on Learning Representations (ICLR)},
  2018.

\bibitem{lalonde-siggraph-asia-09}
Jean-Fran\c{c}ois Lalonde, Alexei~A. Efros, and Srinivasa~G. Narasimhan.
\newblock Webcam clip art: Appearance and illuminant transfer from time-lapse
  sequences.
\newblock {\em ACM Transactions on Graphics (SIGGRAPH Asia)}, 2009.

\bibitem{larsson2019cross}
Mans Larsson, Erik Stenborg, Lars Hammarstrand, Marc Pollefeys, Torsten
  Sattler, and Fredrik Kahl.
\newblock A cross-season correspondence dataset for robust semantic
  segmentation.
\newblock In {\em Conference on Computer Vision and Pattern Recognition
  (CVPR)}, 2019.

\bibitem{lowe1999object}
David~G Lowe.
\newblock Object recognition from local scale-invariant features.
\newblock In {\em International Conference on Computer Vision (ICCV}, 1999.

\bibitem{RobotCarDatasetIJRR}
Will Maddern, Geoff Pascoe, Chris Linegar, and Paul Newman.
\newblock {1 Year, 1000km: The Oxford RobotCar Dataset}.
\newblock {\em The International Journal of Robotics Research (IJRR)}, 2017.

\bibitem{maddern2014illumination}
Will Maddern, Alex Stewart, Colin McManus, Ben Upcroft, Winston Churchill, and
  Paul Newman.
\newblock Illumination invariant imaging: Applications in robust vision-based
  localisation, mapping and classification for autonomous vehicles.
\newblock In {\em International Conference on Robotics and Automation (ICRA)},
  2014.

\bibitem{min2009deep}
Renqiang Min, David~A Stanley, Zineng Yuan, Anthony Bonner, and Zhaolei Zhang.
\newblock A deep non-linear feature mapping for large-margin knn
  classification.
\newblock In {\em International Conference on Data Mining (ICDM)}, 2009.

\bibitem{olid2018single}
Daniel Olid, José~M. Fácil, and Javier Civera.
\newblock Single-view place recognition under seasonal changes.
\newblock In {\em IROS Workshop on Planning, Perception, Navigation for
  Intelligent Vehicle (PPNIV)}, 2018.

\bibitem{qu2017deshadownet}
Liangqiong Qu, Jiandong Tian, Shengfeng He, Yandong Tang, and Rynson~WH Lau.
\newblock Deshadownet: A multi-context embedding deep network for shadow
  removal.
\newblock In {\em Conference on Computer Vision and Pattern Recognition
  (CVPR)}, 2017.

\bibitem{robertini2018illumination}
Nadia Robertini, Florian Bernard, Weipeng Xu, and Christian Theobalt.
\newblock Illumination-invariant robust multiview 3d human motion capture.
\newblock In {\em Winter Conference on Applications of Computer Vision (WACV)},
  2018.

\bibitem{ruiz2018deep}
Ariel Ruiz-Garcia, Vasile Palade, Mark Elshaw, and Ibrahim Almakky.
\newblock Deep learning for illumination invariant facial expression
  recognition.
\newblock In {\em International Joint Conference on Neural Networks (IJCNN)},
  2018.

\bibitem{sandler2018mobilenetv2}
Mark Sandler, Andrew Howard, Menglong Zhu, Andrey Zhmoginov, and Liang-Chieh
  Chen.
\newblock Mobilenetv2: Inverted residuals and linear bottlenecks.
\newblock In {\em Conference on Computer Vision and Pattern Recognition
  (CVPR)}, 2018.

\bibitem{shu2017portrait}
Zhixin Shu, Sunil Hadap, Eli Shechtman, Kalyan Sunkavalli, Sylvain Paris, and
  Dimitris Samaras.
\newblock Portrait lighting transfer using a mass transport approach.
\newblock {\em ACM Transactions on Graphics (TOG)}, 2017.

\bibitem{simonyan2014very}
Karen Simonyan and Andrew Zisserman.
\newblock Very deep convolutional networks for large-scale image recognition.
\newblock {\em arXiv preprint arXiv:1409.1556}, 2014.

\bibitem{sun2019single}
Tiancheng Sun, Jonathan~T Barron, Yun-Ta Tsai, Zexiang Xu, Xueming Yu, Graham
  Fyffe, Christoph Rhemann, Jay Busch, Paul Debevec, and Ravi Ramamoorthi.
\newblock Single image portrait relighting.
\newblock {\em ACM Transactions on Graphics (TOG)}, 2019.

\bibitem{wang2018stacked}
Jifeng Wang, Xiang Li, and Jian Yang.
\newblock Stacked conditional generative adversarial networks for jointly
  learning shadow detection and shadow removal.
\newblock In {\em Conference on Computer Vision and Pattern Recognition
  (CVPR)}, 2018.

\bibitem{zhang2020portrait}
Xuaner Zhang, Jonathan~T. Barron, Yun-Ta Tsai, Rohit Pandey, Xiuming Zhang, Ren
  Ng, and David~E. Jacobs.
\newblock Portrait shadow manipulation.
\newblock In {\em ACM Transactions on Graphics (TOG)}, 2020.

\bibitem{zhang2015learning}
Xi Zhang, Yanwei Fu, Andi Zang, Leonid Sigal, and Gady Agam.
\newblock Learning classifiers from synthetic data using a multichannel
  autoencoder.
\newblock {\em arXiv preprint arXiv:1503.03163}, 2015.

\bibitem{zhou2019deep}
Hao Zhou, Sunil Hadap, Kalyan Sunkavalli, and David~W Jacobs.
\newblock Deep single-image portrait relighting.
\newblock In {\em International Conference on Computer Vision (ICCV)}, 2019.

\end{thebibliography}
}

\end{document}